\documentclass{article}

\usepackage[utf8]{inputenc}

\title{Non-Determinism in TensorFlow ResNets}
\author{Miguel Morin, Matthew Willetts}
\date{January 2020}

\usepackage{graphicx}
\usepackage{hyperref}
\usepackage{booktabs}
\usepackage[margin=0.95in]{geometry}
\begin{document}

\maketitle

\begin{abstract}
    We show that the stochasticity in training ResNets for image classification on GPUs in TensorFlow is dominated by the non-determinism from GPUs, rather than by the initialisation of the weights and biases of the network or by the sequence of minibatches given. The standard deviation of test set accuracy is 0.02 with fixed seeds, compared to 0.027 with different seeds---nearly 74\% of the standard deviation of a ResNet model is non-deterministic.
    For test set loss the ratio of standard deviations is more than 80\%.
These results call for more robust evaluation strategies of deep learning models, as a significant amount of the variation in results across runs can arise simply from GPU randomness.
\end{abstract}

\section{Introduction}

Commonly researchers need to run deep learning models repeatedly to understand the variation in performance.
The re-runs will commonly be done using new seeds for the creation of minibatches and for initialisation.

We show here that for ResNets in TensorFlow + Keras running on GPUs the variation caused by these sources of noise is dominated by that coming from the intrinsic non-determinism of GPUs themselves.
While the existence of GPU non-determinism is well-known, the scale of its effect is perhaps less well understood -- especially in the context of contemporary machine learning algorithms.

First we explain the source of this GPU-induced variability, across packages and operating systems.
Then we study the effects of GPU non-determinism on standard ResNet architectures.
To isolate the effect of GPU non-determinism we held constant the sources of randomness that effect the training of a ResNet other than GPU non-determinism: initial parameter weights and the ordering of training minibatches.

\section{GPU Non-Determinism from Float Operations}

We obtained reproducibility of random number generators by setting three seeds: TensorFlow's graph-level seed, TensorFlow's operation-level seed, and NumPy's seed.
We also reset the GPU with \verb|sudo nvidia-smi -rac| (see deployment code\footnote{\href{https://github.com/miguelmorin/reproducibility/blob/master/deploy_azure.sh}{https://github.com/miguelmorin/reproducibility/blob/master/deploy\_azure.sh}}). We experiment using TensorFlow 1 and 2 running on a local macOS computer, Google Colab (web-based), and Linux (cloud computing on Azure).
We obtained reproducibility of random numbers:

\begin{itemize}
    \item without the operation-level seed, the first random number drawn was reproducible across platforms separately for eager mode and for non-eager mode, but not between these two modes of operation.
  \item with the operation-level seed, the first random number drawn was reproducible across platforms and eager/non-eager modes.
  \end{itemize}

We computed the mean and standard deviation of 100 random numbers.
The results were sometimes different across versions of TensorFlow, across platforms, across modes (eager and non-eager), and between GPU versus CPU.
On GPUs, even with the same software, mode of operation, and platform, the standard deviation varied across runs.
(See the Python script with tests\footnote{\href{https://github.com/miguelmorin/reproducibility/blob/master/reproducibility_tests.py}{https://github.com/miguelmorin/reproducibility/blob/master/reproducibility\_tests.py}} and the full details of the results\footnote{\href{https://github.com/miguelmorin/reproducibility/blob/master/tensorflow-reproducibility.md}{https://github.com/miguelmorin/reproducibility/blob/master/tensorflow-reproducibility.md}}.)

We believe that this variability is due to loss of floating point determinism when performing calculations.
Consider adding these floats: ${2^{-N}, 2, -2^{-N}}$, where the exact result is 2.
If we add them in the order given and $N$ is large enough, the precision of the first number is truncated when adding the second.
Different compilers and architectures sum numbers in different orders, which explains the variability in the mean and standard deviation across systems.

GPUs are non-deterministic because they fill the registers with available data: if $2^{-N}$ is aligned with $-2^{-N}$ and they are added first, we would get the exact result; but if $2^{-N}$ is aligned with $2$, we would get an approximate result.
This explains why the standard deviation is different across runs on GPUs on the same system and with the same seeds.

We would probably obtain reproducibility if we could flash the GPU and reset it to a known state.
This would require owning the full graphics card. On cloud compute platforms, the virtual machines often share such cards.
Thus for cloud compute users this fix is not available.

\subsection{Variability in a deep learning model}

This non-determinism in GPUs and resulting lack of reproducibility are important as deep learning models are highly non-convex, with many local minima.
So a small difference in the result of a computation can propagate and lead the optimisation to very different end states.

We measured the impact of non-determinism with a ResNet-50 model \cite{ResNets} on CIFAR-10 \cite{Krizhevsky2009}.
We trained for 200 epochs with a batch size of 32.
We used the standard reference implementation from Keras \cite{chollet2015keras, tensorflow}\footnote{\href{https://keras.io/examples/cifar10_resnet}{https://keras.io/examples/cifar10\_resnet}} with the minor tweaks of: 1) changing the kernel initialiser of the ResNet layers to have a hardcoded seed of 0 and 2) setting the NumPy and Tensorflow seeds as an argument.
And so we trained the model 50 times with the same NumPy and Tensorflow seeds and 50 times with both seeds set to the number of the run, from 0 to 49.
The accuracy and loss of each are in our repository with name \verb|resnet_different_seeds.csv| and \verb|resnet_same_seed.csv|.

The series of minibatches is deterministic when the system seeds are set as described above, as is the initialisation of the parameters of the model.
%

We then coded the comparison of models by testing the same number of layers and the same array of weights in each layer.

\section{Experiments}

We experiment on ResNets using TensorFlow 1 on Linux on Microsoft Azure.
The code is at our repository.\footnote{
\href{https://github.com/miguelmorin/reproducibility}{https://github.com/miguelmorin/reproducibility}}
We used an Azure NV6 virtual machine, which provides an NVIDIA M60 GPU.

\subsection{Controlling for other sources of randomness}

The final trained model we get depends on the initial values of its parameters and the particular ordering of training data into minibatches.
These procedures use pseudo-random number generators - we sample the initial values of neural network weights from standard probability distributions, and we perform permutations of our data to get our sequence of batches.

If there were no other sources of randomness, then models trained with the same initial values and with the same series of minibatches would be identical.
And of course when training with different seeds for initialisation and permutation the initial models and the training batches are different, so we end up with different trained models.

However, GPU non-determinism is an extra source of randomness.
Even with the same seeds for initialisation and batching we end up with different trained models.
So to isolate this effect, we fixed the seeds for initialisation and batching, and verified that this does result in reproducible values for these processes.


For more details on the equality comparison, see the section "Equality comparison" in the repository \verb|README|\footnote{\href{https://github.com/miguelmorin/reproducibility/blob/master/README.md}{https://github.com/miguelmorin/reproducibility/blob/master/README.md}}.


\subsection{Results}

\begin{table}[h!]
\caption{Standard Deviation of Test Set Accuracy and Loss when training ResNet-50 50 times with the same seed and with different seeds.}
\vspace{3mm}
\centering
\begin{tabular}[pos]{r|c|c}
& $\sigma$(accuracy) & $\sigma$(loss) \\
\hline
Same seed & $1.995\times10^{-2}$  & $3.020\times10^{-3}$ \\
\hline
Different seeds & $2.699\times10^{-2}$ & $3.464\times10^{-3}$\\
\hline
\hline
Ratio & 0.739 & 0.872\\
\hline
\end{tabular}
\end{table}


A calculation of standard deviation on the accuracy and loss in the table shows striking results: the variation from non-determinism in GPU calculations accounts for between 74\% and 87\% of the standard deviation in the scores of a CIFAR neural network.


\section{Conclusion}

These results pose a challenge to deep learning research.
Models are often assessed by the improvement in accuracy relative to a standard, but our paper shows that around 80\% of the standard deviation in the accuracy of a ResNet model is non-deterministic, much greater than that associated with the other sources of randomness in training.
We propose that the literature move to using distributions in the assessment of the improvement of a new model: instead of comparing the accuracy of a new model to the accuracy of a standard model, which may seem favourable simply by luck, we should be comparing the distribution of accuracy between the new and the standard model.
This recommendation is in line with Machine Learning Reproducibility Checklist\cite{reprod}, now used by NeurIPS, that asks researchers to give both error bars and a measure of variation for results.

Attempting to isolate the variation due to non-determinism in GPU training will allow a comparison of distributions and a statistical test of whether a model is better than a given baseline, and also allow a better quantification of the improvement not only in the average performance but also its variation.
After all, a model that trains with lower variance is worthwhile in itself.

These preliminary results are only for ResNets. We hope to stimulate interest in the effects of GPU non-determinism in other model classes.

\newpage
\bibliographystyle{unsrt}
\bibliography{references}
\end{document}